# PACTHub: a large-scale structured experimental dataset for three-dimensional photoacoustic computed tomography

Zhibo Xiao[a,b], Xuanhao Wang[b,c,*], Hao Huang[a,b], Fan Meng[a,b], Xuran Zhou[a,b], Xiali Gao[a,b], Yang Xiao[b], Junhui Shi[a,b,c,*]

[a] College of Biomedical Engineering & Instrument Science, Zhejiang University, Hangzhou 310027, China

[b] Zhejiang Lab, Hangzhou 311100, China

[c] State Key Laboratory of Ocean Sensing, Hangzhou 311200, China

* Corresponding authors: Xuanhao Wang (xh-wang@zhejianglab.org); Junhui Shi (junhuishi@outlook.com).

## Abstract

Three-dimensional photoacoustic computed tomography (PACT) provides volumetric contrast related to optical absorption for studying tissue anatomy and dynamics. Method development requires experimental datasets that retain measured signals and acquisition context. Public resources remain limited in their combined coverage of varied specimens, volumetric acquisition modes and reusable reconstruction information. Here we present PACTHub, a large-scale structured experimental dataset comprising 1,464 acquisition groups and 99,810 photoacoustic exposures collected over nearly four years. Animal, phantom and human measurements span single-position imaging, rotational compounding, panoramic scanning and time series. A common PAdata/JSON representation links acoustic signals to detector geometry, exposure selection, signal preparation and physical reconstruction coordinates. Public reconstruction implementations, numerical volumes and maximum-intensity projections provide practical entry points for reusing the measurements. Spatial and temporal examples demonstrate how retained acquisition relationships support panoramic reconstruction and regional signal analysis. Detector and exposure subsets enable studies of angular sampling and measurement budget. Volumetric evaluation combines PSNR, SSIM and L1 with Q, a complementary measure of local-contrast agreement within reference-defined structures. By connecting measured signals to reconstruction settings and numerical outputs, PACTHub makes its experimental scale and diversity available for reuse. It provides an experimental foundation for developing and evaluating artificial intelligence (AI) methods in volumetric PACT.

## 1 Introduction

Three-dimensional photoacoustic computed tomography (PACT) provides volumetric images of optical-absorption-related contrast for studying tissue anatomy and dynamics. High-speed and panoramic systems now support volumetric brain and breast imaging, extended anatomical coverage and repeated observations over time [1–3]. The information recovered from these acquisitions depends on how signals are sampled and combined, as well as on the reconstruction and analysis methods applied to them. Shared experimental datasets allow these choices to be investigated under common measurement conditions. They also make method development accessible beyond the laboratories that acquire the data.

Experimental measurements capture the combined effects of illumination, detector response, acoustic propagation, motion and noise. Numerical phantoms and virtual imaging frameworks provide controlled conditions and known targets for studying PACT [4–10]; measured data complement them by allowing methods to be examined under actual acquisition conditions. In volumetric PACT, detector coverage and the combination of multiple exposures influence the recovered structures. Retaining individual exposures allows different detector selections and measurement budgets to be investigated using the same acquisition. Preserving relationships among scan positions and time points similarly allows spatial reconstruction and temporal analysis to be revisited. These uses require access to the measurements underlying the final images.

Public data resources have established a basis for shared method development. OADAT links experimental forearm data to defined processing tasks [11]. Quantitative phantom resources, including OpenQPAIData, provide characterized material properties [12,13], while digital twins supply model-derived references for reconstruction assessment [14]. Expert image-quality ratings have also been shared through PhotIQA [15]. Volumetric measurements and reconstruction products have been released alongside three-dimensional imaging studies [1,2,16–18]. These volumetric releases are organized around particular imaging studies and differ in the measurements and accompanying reconstruction information they provide. There remains a need for large experimental three-dimensional PACT collections spanning varied specimen categories and spatial or temporal acquisition modes. Such collections require sufficient physical context to reconstruct and reanalyze the measurements.

For such collections, acquisition scale and structured organization serve complementary purposes. Coverage across specimen categories and acquisition modes provides measured variation, while repeated exposures support comparisons within an acquisition. Reuse of each measurement depends on its association with detector geometry, timing and reconstruction settings, as reflected in the IPASC approach to storing photoacoustic signals and metadata [19]. Reusable analysis tools and the FAIR principles provide further foundations for computational access to research data [20,21]. Preserving these associations across acquisitions makes it possible to prepare consistent inputs, change reconstruction choices and interpret numerical outputs under documented conditions.

These data are also important for artificial intelligence (AI), including model-based learned reconstruction and sparse-data image enhancement in PACT [22,23]. In photoacoustic artifact compensation, learning-curve experiments show that training-set size affects image-recovery quality [24]. Chest radiograph classification experiments similarly show improved performance with more training data, with diminishing gains at the larger tested sizes [25]. Data composition also matters: selecting less redundant images can improve data efficiency in medical imaging [26]. Studies of photoacoustic oximetry show that simulated training distributions affect estimation accuracy on target data [27]. Together, these findings motivate experimental PACT resources that combine acquisition scale with coverage of relevant specimen and measurement conditions. Linking signals, reconstruction settings and image products makes this variation accessible for preparing and evaluating learning tasks.

Here we present PACTHub, a large-scale structured experimental dataset comprising 1,464 acquisition groups and 99,810 photoacoustic exposures collected over nearly four years. Animal, phantom and human measurements span single-position imaging, rotational compounding, panoramic scanning and time series. The dataset associates measured signals with their physical acquisition and reconstruction settings, together with numerical image products where available. We characterize the collection and demonstrate how its retained spatial, temporal and sampling relationships support reconstruction and analysis. Public reconstruction implementations and complementary volumetric evaluation measures provide practical means of using the data. By linking this experimental coverage to reusable reconstruction information, PACTHub provides a shared data foundation for method development and AI research in three-dimensional PACT.

## 2 Materials and methods

### 2.1 Data acquisition and structured organization

An acquisition group denotes a measurement record before scan stitching, and an exposure denotes one photoacoustic excitation together with its multichannel signals. Five previously verified duplicate waveform copies are excluded and documented in the inventory and exclusion list. Each release-relative acquisition-directory path is a unique record key. Original timestamps remain searchable identifiers, but can recur across source collections. Records retain known relationships among scan positions, temporal acquisitions and duplicate groups. Repeated groups may originate from the same specimen; subject independence cannot be verified across the archive. The reported totals describe acquisition records rather than independent subjects.

The experimental acquisition setup in Fig. 1a generated the multichannel signals represented by the PACTHub records. It uses a 1,024-element hemispherical array for animal, phantom and human imaging. Stationary acquisition, rotational compounding and translation across scan positions provide the acquisition modes in Fig. 1b. Temporal experiments retain acquisition order and elapsed time. The imaging systems have been described previously [2,3]; the PACTHub inventory and acquisition records define the collection summarized here.

The PAdata and associated reconstruction records for the acquisitions summarized in Fig. 1 are available through Google Drive. Numerical volumes and physical axes support volumetric analysis, while XY maximum-intensity projections (MIPs) provide previews. An interactive visual preview of the dataset is available at https://pacthub.net/, and a representative subset is archived on Zenodo. Acquisition-specific geometry is retained, including registered detector positions and receive directions for the human-hand protocol.

A standardized sample directory associates PAdata with reconstruction_parameters.json and sample_metadata.json. HDF5-backed MAT signals use frame–time–channel storage, corresponding to a channel–time–frame view in MATLAB. Numerical volumes use Z–Y–X order, and spatial coordinates are expressed in millimetres. The records specify array keys, dimensions, data types, amplitude scaling and channel-mapping states. Relative paths are resolved from the parameter-file directory. A prepared-profile JSON Schema and field dictionary describe the input structure in Supplementary Methods S1 and the accompanying source files. The loader performs additional checks on file contents and execution parameters.

The reconstruction record groups the physical and computational settings by function (Fig. 2a). The signal section describes sampling, timing, filtering and amplitude preparation; frames specifies exposure selection and compounding; grid defines the physical field and spatial sampling; and geometry defines detector positions and acquisition-dependent transformations. Reconstruction and model-based (MB) sections contain algorithm settings. Detector coordinates can be embedded as detector_xyz_mm, so an embedded profile needs no separate detector-coordinate parameter file. Protocol-specific transformations remain part of the record. Grid sampling rules and derived physical axes are retained explicitly, including the distinction between round(length × pixels/mm) points and the interval-based rule that adds one endpoint.

Reconstruction proceeds by validating a sample, preparing an effective profile and running the selected implementation. Validation checks signal dimensions, exposure indices, geometry and required settings. Preparation resolves available source parameters, embeds detector coordinates when requested, and records the supplied or selected field of view (FOV) and sound speed in a new profile. A profile with unresolved required parameters is not executed by substituting guessed physical values. The backprojection (BP) and MB reconstruction routes export numerical volumes, physical axes, MIPs and execution records. Supplementary Methods S1 describes the commands and field conventions; each acquisition retains its own numerical settings.

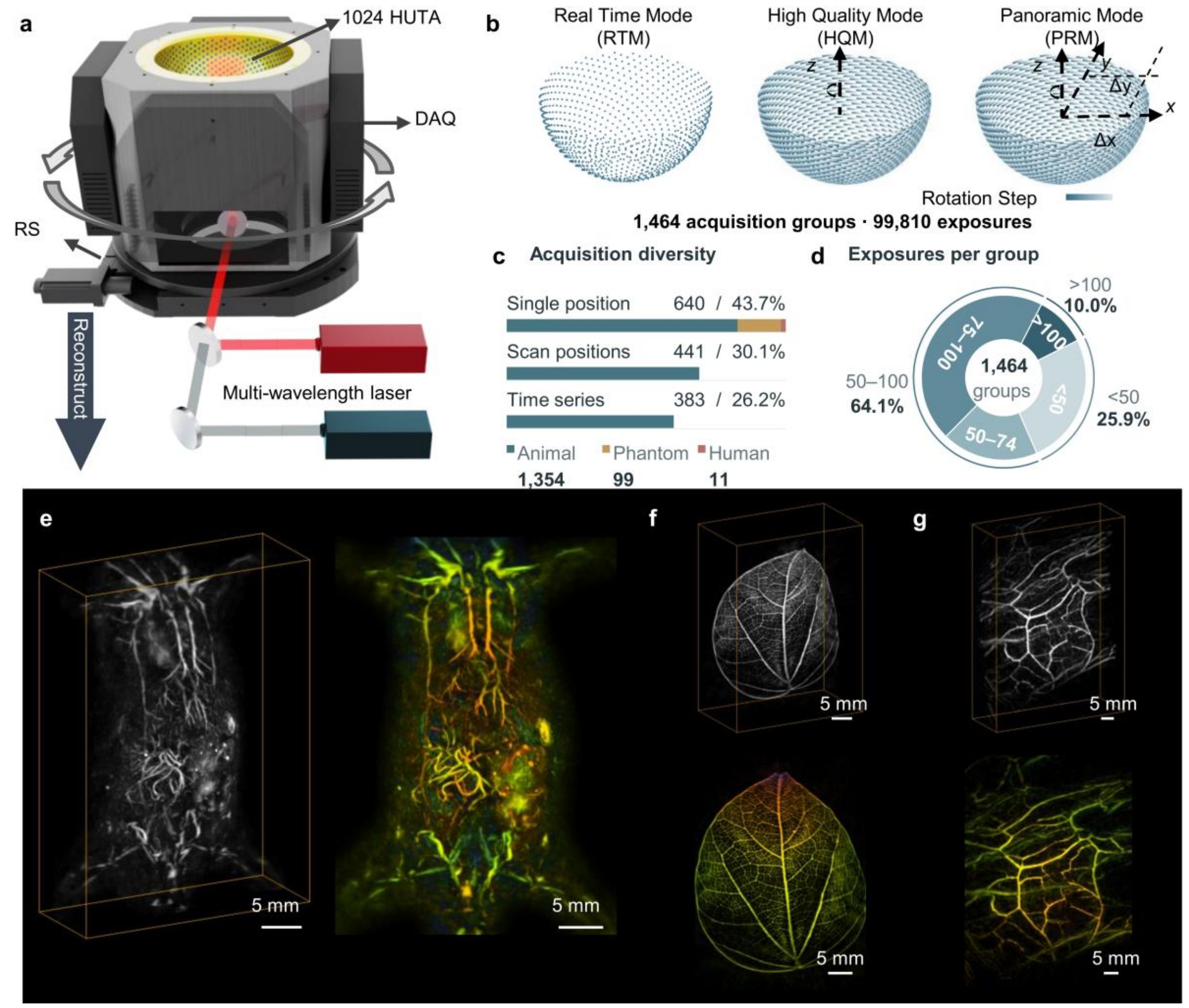

**Figure 1. Experimental acquisition and composition of PACTHub.** (a) Experimental setup used to acquire the recorded signals. (b) Stationary, rotational and panoramic acquisition modes. (c) Specimen and acquisition composition. (d) Exposures per acquisition group. (e–g) Animal, leaf and hand volumes shown as grayscale renderings and depth-encoded MIPs. The animal volume is from 3D-PanoPACT [2]. Brightness is adjusted independently by specimen. Scale bars, 5 mm.

## 2.2 Reconstruction workflow

Reconstruction begins with two sequential parameter-selection procedures (Fig. 2b,c): identifying the effective signal region to define the FOV, then selecting effective sound speed within that fixed field. FOV selection starts from a coarse BP volume over an enlarged search region. Gaussian scales of 0.55 and 6 mm convert its XY MIP to positive local contrast. A threshold combining Otsu separation with a median-plus-three-MAD criterion identifies strong signal regions. Morphological closing, hole filling and connected-region filtering consolidate the support, and a vessel-continuity prior adds weaker connected structures. Principal-component analysis estimates the support's long axis and oriented bounds. The estimated XY center and rotation update the reconstruction coordinates. The Z center and target depth remain supplied parameters because the localization projection determines XY support.

Effective sound speed is selected after fixing the exposures, signal processing and final FOV. Image-based autofocus is an established approach to this parameter choice [28]. In Fig. 2c, the same 20 exposures are reconstructed at 23 candidate speeds from 1488 to 1499 m/s in 0.5 m/s increments. A common support Ω is formed from the median of percentile-normalized candidate MIPs, using a threshold of 0.08, morphological opening and closing, and removal of four-pixel borders. Each numerical MIP is smoothed with σ = 1 pixel to obtain $I_c$. The focus score is

$$J(c) = \frac{\langle |\nabla I_c|^2 \rangle_\Omega}{Var_\Omega(I_c)} \tag{1}$$

The selected speed maximizes J(c). Scores are unevaluable if the support contains fewer than 32 pixels or the variance of the smoothed intensity is at most $10^{-24}$. All candidates use the same support, and the four displayed details share their coordinates and brightness window. The score is calculated from the complete numerical MIP within Ω. The selected value is an effective homogeneous propagation parameter for the reconstruction model. Support masks, candidate scores and the chosen value are retained with the reconstruction settings.

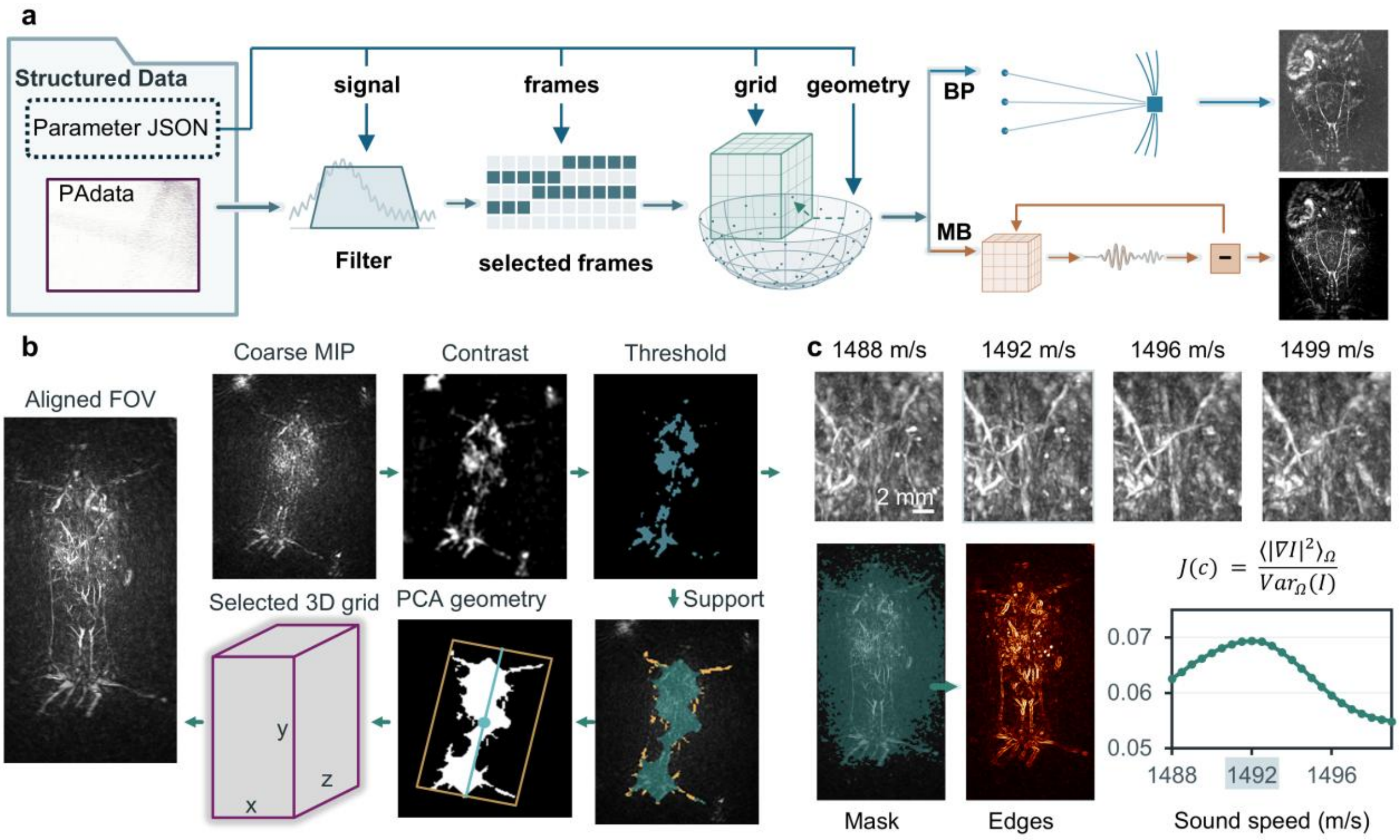

**Figure 2. Structured inputs and reconstruction settings.** (a) PAdata and JSON records linked to the BP/MB workflow. (b) FOV selection from coarse reconstruction, segmented support and fitted geometry. Teal marks accepted support; amber marks vessel-continuity additions. (c) Four matched sound-speed reconstructions, the common support, gradient maps and focus curve. Scale bar, 2 mm.

With the field and sound speed specified, the BP and MB routes reconstruct the selected exposures using the shared measurement records and physical coordinates. The public BP implementation follows the universal backprojection formulation of Xu and Wang [29], using $2(p - t\partial p/\partial t)$, linear interpolation in time and solid-angle weighting, with CPU and GPU execution. Supplementary Methods S2 specifies the public UBP execution protocol and two-file parameter profile for leaf acquisition 20241115_104329. The stored BP comparison volumes in the main figures were generated with a separate confidential in-house implementation. Its weighted BP computation is described by the pseudocode in Supplementary Methods S6; the public UBP is a distinct reconstruction operator with the same data interface.

Model-based acoustic inversion estimates an initial pressure distribution using a forward model of acoustic propagation [30]. The supplied MB implementation uses finite-disk forward and adjoint operators and a nonnegative iterative solver with L1, three-dimensional total-variation and L2 penalties. The stored mouse-comparison outputs use zero initialization and five fixed iterations, a 1.75 mm receiver radius, 48 quadrature points per detector, and a 129-sample temporal kernel. The first 2,000 signal samples are masked and the recorded predelay is −42 samples. The one- and five-exposure runs retain their original regularization coefficients and primal/dual steps. Supplementary Methods S5 reports these settings, objective trajectories and runtime. The comparison uses the stored volumes with their branch-specific preprocessing.

### 2.3 Spatial and temporal analysis

Representative renderings and depth-encoded MIPs are generated from numerical volumes. The animal overview in Fig. 1 uses the author-released processed 3D-PanoPACT volume, with its source explicitly attributed; the leaf and hand use their corresponding reconstructions. Grayscale renderings represent intensity, and depth encoding assigns color from the position contributing to the projection. Source attribution and display settings are retained for each visualization.

The spatial example uses acquisition 20260114_192341_Scan1 to relate one exposure, rotational compounding and scan-plus-rotation imaging (Fig. 5). The single-exposure and rotation views use the same central scan position. Twenty selected exposures are retained at each of 70 positions arranged on a 7 × 10 trajectory with 6 × 8 mm steps. Each stage combines two BP volumes by a voxelwise unit-weight sum. Their fourth-order zero-phase Butterworth passbands are 1.44–5.76 MHz and 0.5–8 MHz; both use 1508.5 m/s, a −41-sample onset delay and 0.1 mm spatial sampling. The scan is fused using Gaussian and frame-count weighting with a 30 mm full width at half maximum.

All three views use the physical display field x = −14 to 50 mm, y = −84 to 21 mm and z = 1.9 to 21.9 mm. Recorded display rotations align the long axis by 6° counterclockwise for the single-position views and 2° for the fused scan. The website visualization combines 0.7 independently normalized BP with 0.3 independently normalized Frangi response, without additional fluence gain, and applies the recorded linear brightness window for each stage. The main grayscale panels were verified against the corresponding website arrays. Matched 12 × 12 mm details show the same spatial region. These operations define the visualization; quantitative comparisons use floating-point volumes before display mapping.

Temporal analysis extends this volumetric workflow to repeated measurements in two time series (Fig. 6). The whole-body series retains 42 acquisitions from 20250103_172154 to 20250103_173142, starting at the second acquisition of the original sequence and spanning 9.80 min. Each time point uses the two-bandpass BP summation described above and a fixed effective sound speed of 1500 m/s. The separate abdominal series, acquired at 800 nm, comprises 20 acquisitions from 20260804_110114_800nm to 20260804_112028_800nm over 19.23 min and uses its recorded BP processing at 1499 m/s. Each volume uses 20 selected exposures and a fixed 40 × 80 × 20 mm grid at approximately 0.1 mm sampling. Successive volumes are rigidly registered in the displayed XY coordinates.

Regional measurements use five initial markers per series, each defining a search neighborhood of radius 2 mm. Within each neighborhood, a circular region of interest (ROI) of radius 0.9 mm is selected for coherent temporal variation, subject to a baseline-amplitude floor and spatial regularity. Its XY location is then fixed across time. A reference-defined depth gate of 29 slices, spanning approximately 2.81 mm between slice centers, is also held fixed. For each registered volume, the mean of the depth-restricted MIP within the ROI is divided by the first measured value. Images within a series share a display amplitude window. Because temporal behavior contributes to ROI selection, the analysis is exploratory.

Trend curves are fitted at the actual acquisition times by least squares with a second-derivative penalty and the initial relative value constrained to one. Effective degrees of freedom are seven for the whole-body series and five for the abdominal series; natural cubic interpolation is used for display. All 42 or 20 measured values per ROI are plotted. Shading represents the local minimum–maximum range within adjacent three-acquisition neighborhoods, expanded to include the fitted curve and interpolated with shape-preserving boundaries. It is not a confidence interval. The source records retain acquisition times, measured means, normalized values, ROI coordinates, depth gates and registration transforms. P1–P5 denote selected locations, and relative PA amplitude is interpreted as a signal readout rather than a calibrated physiological quantity.

### 2.4 Benchmark design and evaluation

Numerical assessment combines whole-volume reference agreement with local-contrast agreement on a fixed reference support. Peak signal-to-noise ratio (PSNR) and L1 are calculated over the full three-dimensional field. L1 is the mean absolute voxel difference, and PSNR uses mean squared error with a reference-defined dynamic range. The structural similarity index (SSIM) uses the local luminance, contrast and covariance formulation [31], a three-dimensional Gaussian kernel with $\sigma = 1.5$ voxels, a 13-voxel window, population moments, and constants K1 = 0.01 and K2 = 0.03. Scores are averaged over valid window centers after excluding a six-voxel border. Background remains part of this assessment; the SSIM domain is not restricted to the support used by Q.

In the digital phantom and BP/MB examples, each nonconstant original volume V is independently normalized as

$$V_N = \frac{V - min(V)}{max(V) - min(V)} \tag{2}$$

All four metrics in Tables 1 and 2 use these normalized volumes, a data range of one and gain fixed at one. Constant or nonfinite volumes are rejected. Independent min–max normalization removes positive global gain and offset, so these scores describe normalized reference agreement rather than absolute amplitude calibration. Figure 7 retains the reconstruction scale obtained using the recorded radiofrequency (RF)-derived normalization proxy and exposure averaging, with gain one and a common reference range. The proxy is calculated for each exposure before detector subsampling; Supplementary Methods S4 specifies its definition and signal window. The normalization convention is held fixed within each comparison.

Q provides a complementary readout for structurally sparse three-dimensional PACT volumes with substantial background artifacts. Its workflow in Fig. 3a measures local-contrast agreement within structures defined from a reference volume. For an image or volume I, the positive local contrast is

$$h(I) = max[I - G_\sigma * I, 0], \ \sigma = 0.4\,mm \tag{3}$$

Gσ is a Gaussian kernel with physical scale σ = 0.4 mm. A reference-only mask M combines positive local contrast with a multiscale Hessian tubular response [32], evaluated at 0.20, 0.35 and 0.50 mm after 2 mm background correction. A reference noise proxy is estimated as 1.4826 times the median absolute deviation of signed local contrast among voxels at or below the reference median intensity, with a floor of $10^{-8}$ times the reference maximum. Low/high response thresholds of 0.1/0.3 are paired with contrast thresholds of three/six times this proxy. Retained 26-connected components must contain a high-threshold seed, occupy at least 0.064 $mm^3$ and have a maximum bounding-box extent of at least 1 mm. A 1 mm outer border is excluded. Supplementary Methods S3 specifies Gaussian boundary handling, Hessian normalization and implementation details.

The median positive reference contrast within M divides its voxels into weak and strong groups. The weak group contains h(R) ≤ median, including ties; the strong group contains h(R) > median. These groups remain fixed for all candidates X. The contrast overlap for each group is

$$D_s = \frac{2 \sum_{i \in M_S} min\{h(X)_i, h(R)_i\}}{\sum_{i \in M_S}[h(X)_i + h(R)_i]}, \quad s \in \{weak, strong\} \tag{4}$$

The two overlaps are combined as

$$Q = 100 \sqrt{D_{weak} \, D_{strong}} \tag{5}$$

Each overlap is one minus the summed absolute contrast difference divided by the summed candidate and reference contrast in that group. The geometric mean gives the weak and strong groups equal logarithmic weight. For valid positive reference groups, Q ranges from 0 to 100: identical inputs give 100 and a zero candidate gives zero before independent min–max normalization. Evaluation is invalid when fewer than 100 reference voxels are retained, either group is empty, or metric inputs are nonfinite. Q clips negative local contrast and evaluates agreement within reference support. PSNR, whole-volume SSIM and L1 describe the wider volume, including background and structures outside that support. Figure 3a summarizes the reference-to-contrast-to-grouping workflow. The perturbations and measurement subsets below define the benchmark examples; their results are reported in Section 3.3.

The evaluation measures are first examined using controlled perturbations of a digital phantom (Fig. 3b). The volume contains a strong Gaussian-profile line and two weaker lines on a 64 × 96 × 96 grid at 0.1 mm sampling. Peak amplitudes are 1.00, 0.25 and 0.18, and transverse Gaussian widths are 0.14, 0.10 and 0.09 mm. The background condition adds independent uniform noise in [0, 0.018] only beyond 0.6 mm from every line center, using seed 20261017. Weak-line deletion removes both weaker components; blur uses a Gaussian with σ = 1.5 voxels, equivalent to 0.15 mm. Perturbations are applied before independent volume normalization. The stored arrays, generator and scoring script define the Fig. 3 experiment. Supplementary Methods S3 also includes a false vessel outside the reference support to illustrate the scope of Q.

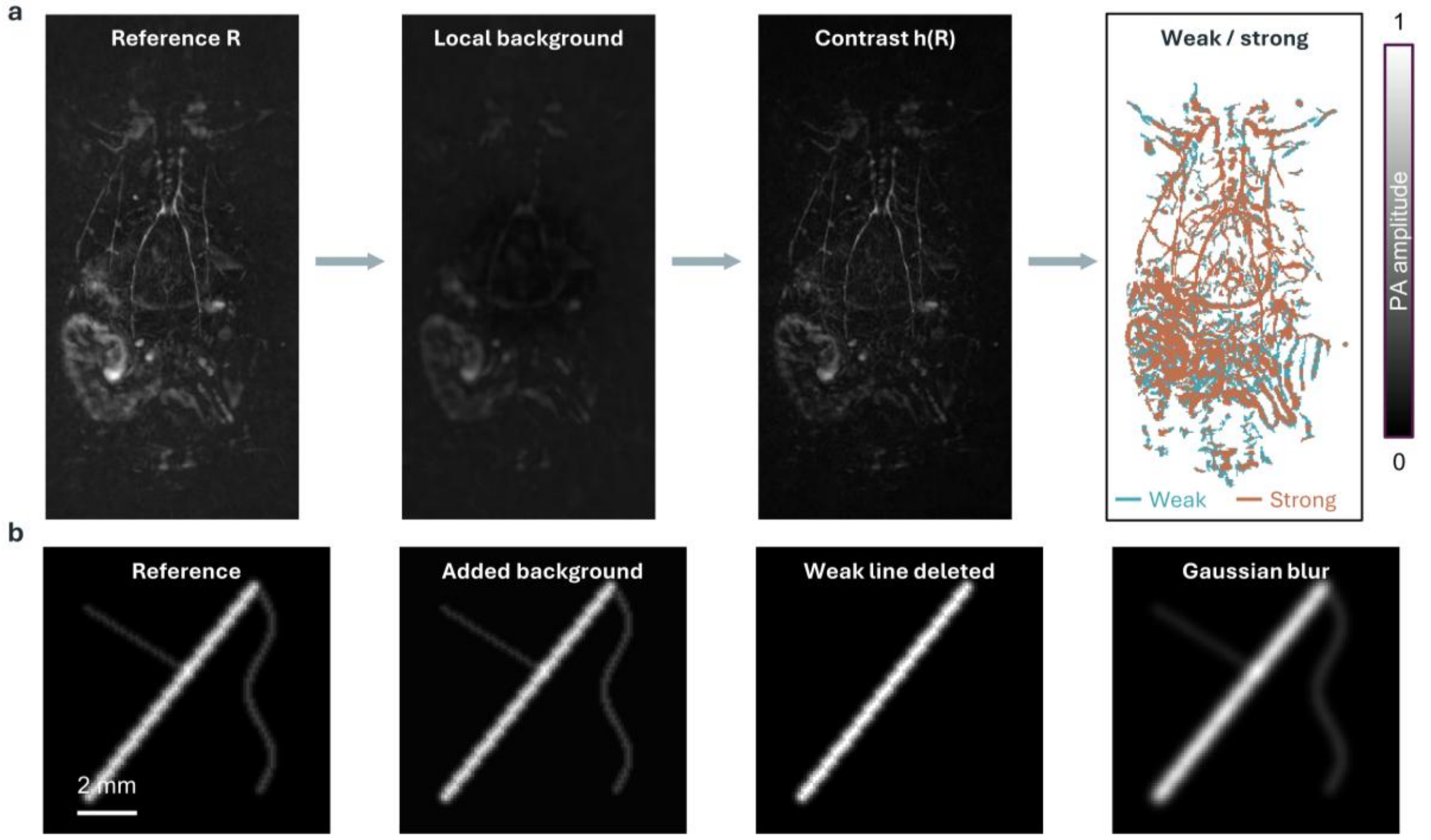

**Figure 3. Q calculation and controlled perturbations.** (a) Mouse reference, local background, positive contrast and reference-defined weak/strong structural groups used by Q. (b) Reference digital line phantom and the added-background, weak-line-deletion and Gaussian-blur conditions. Displayed images are MIPs. Quantitative results are reported in Table 1. Scale bar, 2 mm.

The angular-sampling example uses measured leaf acquisition 20241115_104329 (Fig. 7a). Three disjoint exposure sets define the scoring reference A = {5, 11, 13, 16, 18, 21, 25, 30}, candidates B = {6, 12, 14, 17, 19, 24, 26, 31}, and geometric anchors C = {7, 15, 20, 27}. Each candidate retains 256 channels and eight B exposures. Uniform sampling selects channels across 16 elevation strata; sector sampling selects the 256 detectors nearest a seeded azimuth. The displayed selection uses seed 20261017, with detector IDs recorded in the source data. Volumetric metrics compare candidates with A, while C defines the displayed full-array image and 40 profile anchors. Profiles are interpolated at 0.05 mm intervals, searched within ±0.5 mm and refined by a three-point parabolic peak estimate. Positive boundary maxima are retained; nonpositive peaks receive a 0.5 mm location penalty. Peak Δ is the median absolute peak-location difference from C across all 40 anchors. Supplementary Methods S4 gives distributions and valid-profile counts.

The exposure-budget example constructs nested candidates from the first 1–20 exposures and compares them with a disjoint reference comprising exposures 21–40 (Fig. 7b,c). Geometry, grid, preprocessing and gain remain fixed. Images are displayed for 1, 2, 5, 10, 15 and 20 exposures, while the curves include all twenty counts. These conditions describe changing measurement budget within one acquisition. Their exposure subsets overlap and are statistically dependent.

The stored mouse comparison evaluates one- and five-exposure BP and MB volumes against a 50-exposure BP composite on a common 0.2 mm grid over 40 × 80 × 20 mm (Fig. 8 and Table 2). The single exposure is block 1, and the five-exposure set comprises blocks 1, 13, 26, 39 and 50. BP50 contains these inputs and serves as a higher-exposure reference. All five saved volumes have identical exported axes and shape ZYX [101, 401, 201], using the interval-based endpoint rule. Metric evaluation uses these native arrays without further resampling; the original five-iteration MB outputs are retained. The volumes use the normalization and metrics defined above. Shared inputs, a BP-derived reference and branch-specific preprocessing make this a descriptive comparison of reference agreement; they do not support an independent accuracy ranking of BP and MB.

The examples distinguish acquisition groups, technical exposure subsets and spatial profiles as different analysis units. For learning-based benchmarks, training, validation and test partitions should be grouped by acquisition day. All records from a given day, together with their exposures, scan positions, temporal frames and derived reconstructions, remain in one partition. Known same-specimen or duplicate records spanning different days should also remain together. This grouping prevents related samples from entering both training and evaluation sets; verified subject-disjoint evaluation additionally requires subject identifiers.

## 3 Results

### 3.1 Dataset coverage and reconstruction workflow

PACTHub contains 1,464 acquisition groups and 99,810 exposures acquired from 12 January 2023 to 14 October 2026 (Fig. 1). Animal measurements account for 1,354 groups and 92,580 exposures, phantom measurements for 99 groups and 6,230 exposures, and human measurements for 11 groups and 1,000 exposures. The acquisition organization comprises 640 single-position groups, 441 scan-position groups and 383 time-series groups. These quantities describe acquisition records as defined in Section 2.1, rather than independent subjects.

Six representative three-dimensional views illustrate this diversity (Fig. 4). The scan-plus-rotation mouse captures branching anatomy across the body; the hand, brain and liver show vascular organization at regional and organ scales; and the leaf and square-target phantom provide branching and geometric structures. These views are drawn from the PACTHub sample gallery, where users can compare three-dimensional renderings with grayscale and depth-encoded MIPs.

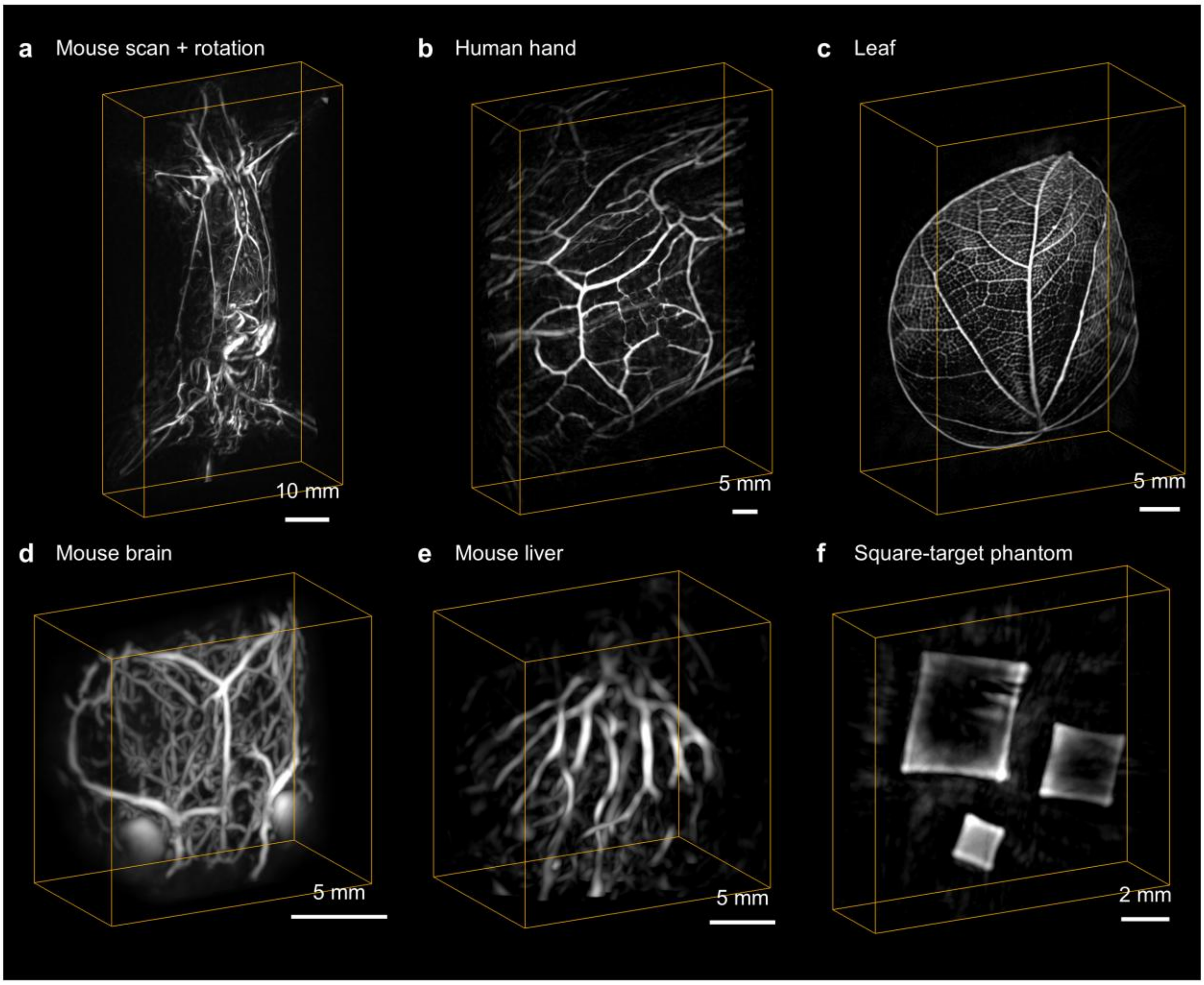

**Figure 4. Representative three-dimensional views from PACTHub.** (a) Mouse scan + rotation. (b) Human hand. (c) Leaf. (d) Mouse brain. (e) Mouse liver. (f) Square-target phantom. Brain and liver views use author-released processed volumes from 3D-PanoPACT [2]. Scale bars, 10 mm (a), 5 mm (b–e) and 2 mm (f).

The archive also retains repeated measurements within acquisitions. Groups with fewer than 50 exposures account for 25.9% of the inventory, those with 50–100 exposures for 64.1%, and those with more than 100 exposures for 10.0%. Exposure selection can therefore be varied while retaining the original acquisition context. Together with scan and temporal relationships, these records provide material for studying measurement budget, view aggregation and repeated observations of the same experiment. Published reference volumes cover 1,315 acquisition groups and are accompanied by 1,525 XY MIP previews: 1,315 GT20 and 210 ScanReference projections. The remaining 149 PA records currently have no published reference volume.

The shared inputs in Fig. 2a make these acquisition records directly usable for numerical reconstruction. The public UBP example reconstructs twenty leaf exposures into a signed 100 × 400 × 200 ZYX volume over 40 × 80 × 20 mm. The recorded RTX 4080 run took 291.1 s including preparation and export, with a maximum CPU–GPU absolute difference of $3.82 \times 10^{-6}$ over 64 checked voxels (Supplementary Methods S2). Numerical volumes and MIPs then provide starting points for reanalysis through the same structured interface.

The parameter-selection example in Fig. 2b,c connects the segmented signal region to the final reconstruction settings. Starting from an 80 × 80 × 20 mm coarse field at 2 voxels/mm, the selected field spans 40 × 80 × 20 mm, with center approximately [0.53, 3.38, −2.00] mm and an additional Z rotation of −12.15°. Within this fixed field, the focus score peaks at 1492 m/s; matched views at 1488, 1492, 1496 and 1499 m/s show the accompanying change in vascular definition. The accepted foreground, fitted axis, bounds, candidate scores and effective parameters are retained with the profile, making the selections available for inspection and reuse.

### 3.2 Spatial and temporal applications

The spatial example retains measurements at individual-exposure, rotational and scan-position levels (Fig. 5). Rotational compounding produces a more continuous branching appearance in the central field, and translation extends the represented anatomy beyond a single position. The panoramic result combines 1,400 selected exposures from 70 positions along the recorded 7 × 10 trajectory. A common physical display field and matched details preserve the spatial correspondence among the three views.

Because individual exposures and scan positions remain identifiable, the same measurement sequence can be revisited at different levels of spatial coverage. Its exposure subsets support studies of measurement reduction and view aggregation, and its position records support panoramic reconstruction and spatial fusion. Figure 5 demonstrates this use of the retained acquisition relationships, with reconstruction, fusion and display settings specified separately.

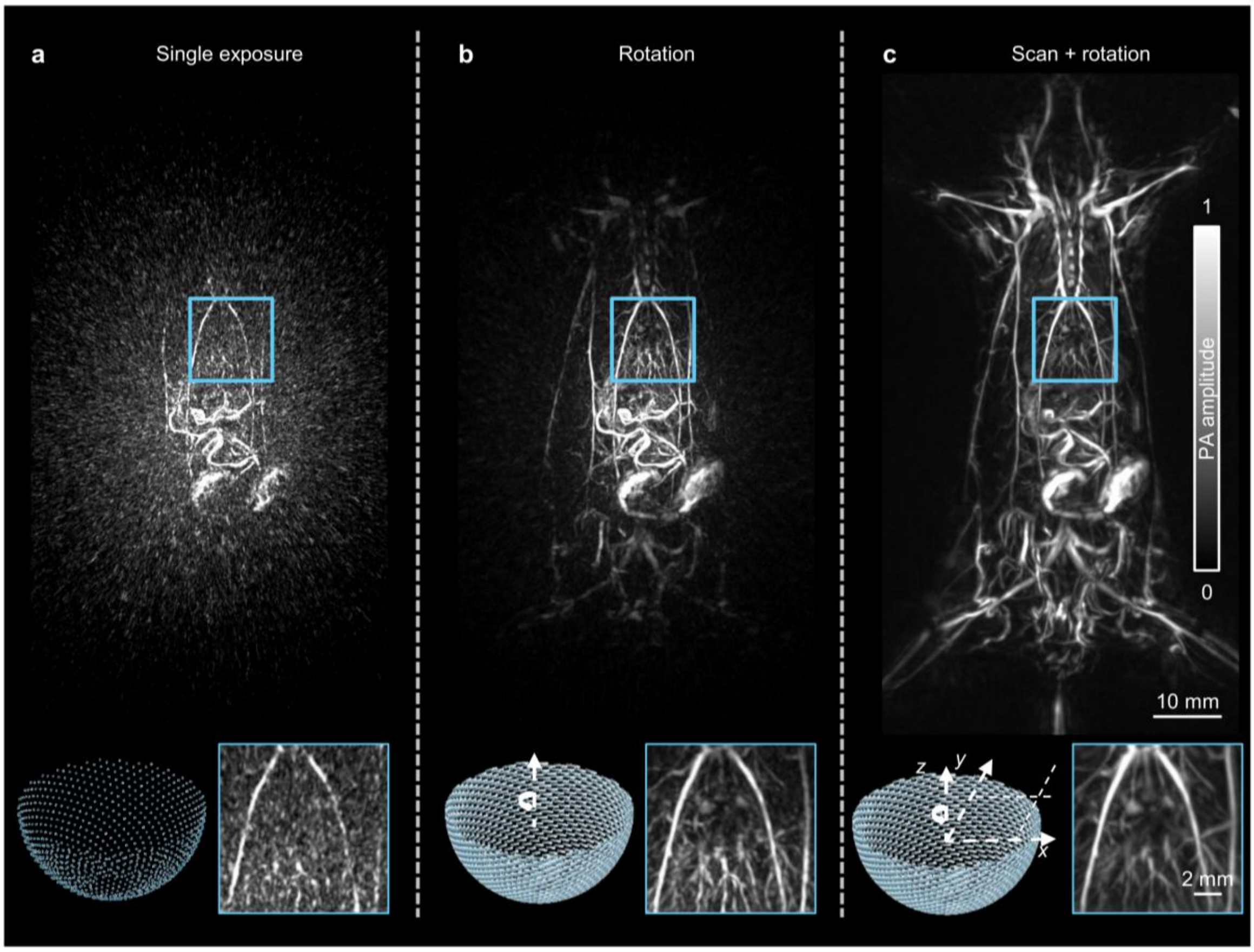

**Figure 5. From single exposure to panoramic reconstruction.** (a) Single exposure. (b) Twenty-exposure rotational compounding. (c) Scan-plus-rotation reconstruction from 70 positions. Schematics show acquisition geometry. MIPs share the physical field and depth interval, with independently set brightness windows. Insets show matched 12 × 12 mm regions. Main and detail scale bars, 10 and 2 mm.

The temporal records support regional tracking across repeated volumes. The whole-body series provides 42 acquisitions over 9.80 min in registered reconstruction coordinates (Fig. 6a). P1 and P2 increase, P3 and P5 finish below their initial amplitudes, and P4 rises before partially declining. At the final measured acquisition, the relative amplitudes of P1 and P2 are 3.74 and 2.77. These values are derived from the measurements. Acquisition identifiers supply the time axis, and fixed ROIs and depth gates define the sampled regions.

The separate 800 nm abdominal series contains 20 measurements over 19.23 min (Fig. 6b). P1 and P2 decline to approximately 0.65 and 0.49 of their initial amplitudes, while P3–P5 increase. Retaining measured points, timestamps and regional definitions allows temporal models and alternative ROIs to be evaluated from the numerical volumes. These exploratory examples demonstrate regional signal analysis across repeated volumes; ROI selection and the interpretation of relative PA amplitude are specified in Section 2.3.

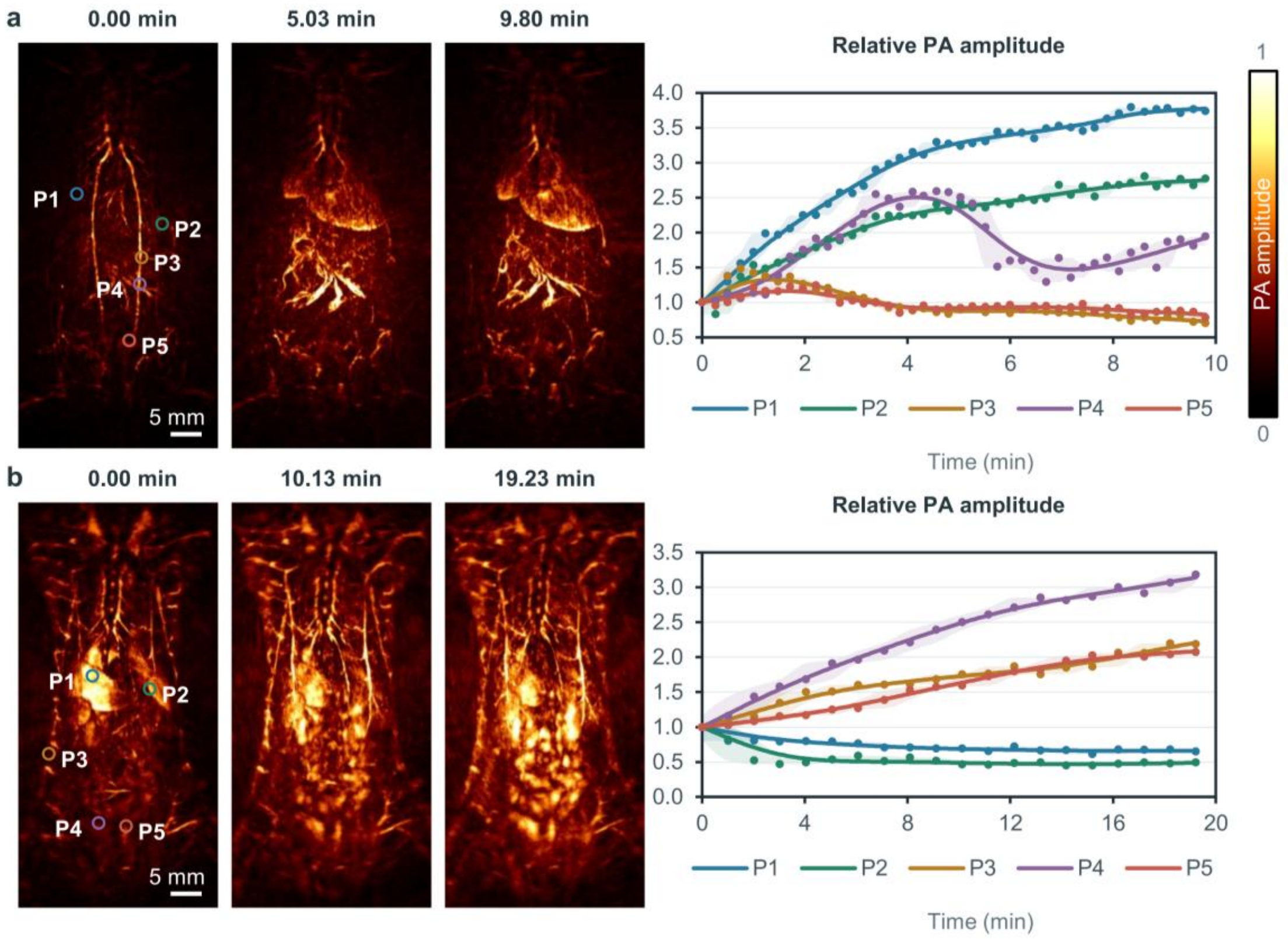

**Figure 6. Regional signal changes in two time series.** (a) Whole-body series with 42 acquisitions over 9.80 min. (b) An 800 nm abdominal series with 20 acquisitions over 19.23 min. P1–P5 mark fixed circular ROIs of radius 0.9 mm. Points show measured relative amplitudes; lines show fitted trends. Shading shows local observed ranges, not confidence intervals. Images share a within-series brightness window.

### 3.3 Reconstruction benchmarking

The benchmark combines controlled perturbations, measured exposure subsets and stored reconstructions to examine image fidelity and structural retention. The controlled perturbations in Fig. 3b first show how Q complements whole-volume metrics under the protocol defined in Section 2.4 (Table 1). Under the stated normalization, added background gives whole-volume SSIM = 0.652 and Q = 97.6. Removing the weak lines gives SSIM = 0.983 and a smaller L1 error, while Q decreases to 66.6. Gaussian blur gives SSIM = 0.960 and Q = 64.6. These controlled changes show that losses of local structural contrast can contribute relatively little to a whole-volume comparison.

**Table 1. Quantitative responses to controlled digital-phantom perturbations.**

| Condition | Q ↑ | PSNR (dB) ↑ | Whole-volume SSIM ↑ | L1 ($\times 10^{-3}$) ↓ |
|---|---|---|---|---|
| Added background | 97.6 | 41.31 | 0.652 | 7.28 |
| Weak-line deletion | 66.6 | 42.62 | 0.983 | 0.631 |
| Gaussian blur | 64.6 | 35.15 | 0.960 | 2.35 |

Reference: the unperturbed digital phantom. Each three-dimensional volume is independently min–max normalized to [0, 1]. PSNR, whole-volume SSIM and L1 include the background; Q uses fixed reference-defined structural support. L1 values are in units of $10^{-3}$. ↑/↓ indicate higher/lower values are better.

Q is therefore reported alongside PSNR, whole-volume SSIM and L1 when inspecting these examples. Its weak and strong groups describe contrast agreement within a fixed reference support, whereas the other measures cover the wider volume. In the additional false-vessel configuration, Q remains approximately 100 while SSIM decreases to 0.980. For this configuration, the added structure has negligible effect on local contrast within the support after normalization. Q provides no direct penalty for structures outside that support, although filtering and normalization can transmit their effects into the scored region. The combined readouts make these different aspects of reconstruction agreement explicit.

The same assessment is then applied to alternative sampling conditions constructed from one measured leaf acquisition using its detector identities and exposure indices (Fig. 7a). At the same budget of 256 channels and eight exposures, uniform sampling better preserves the displayed branching appearance than a contiguous sector. Q is 51.1 for uniform sampling and 37.8 for sector sampling. Median peak-location differences from the separate C-defined anchors are 0.007 and 0.207 mm, respectively.

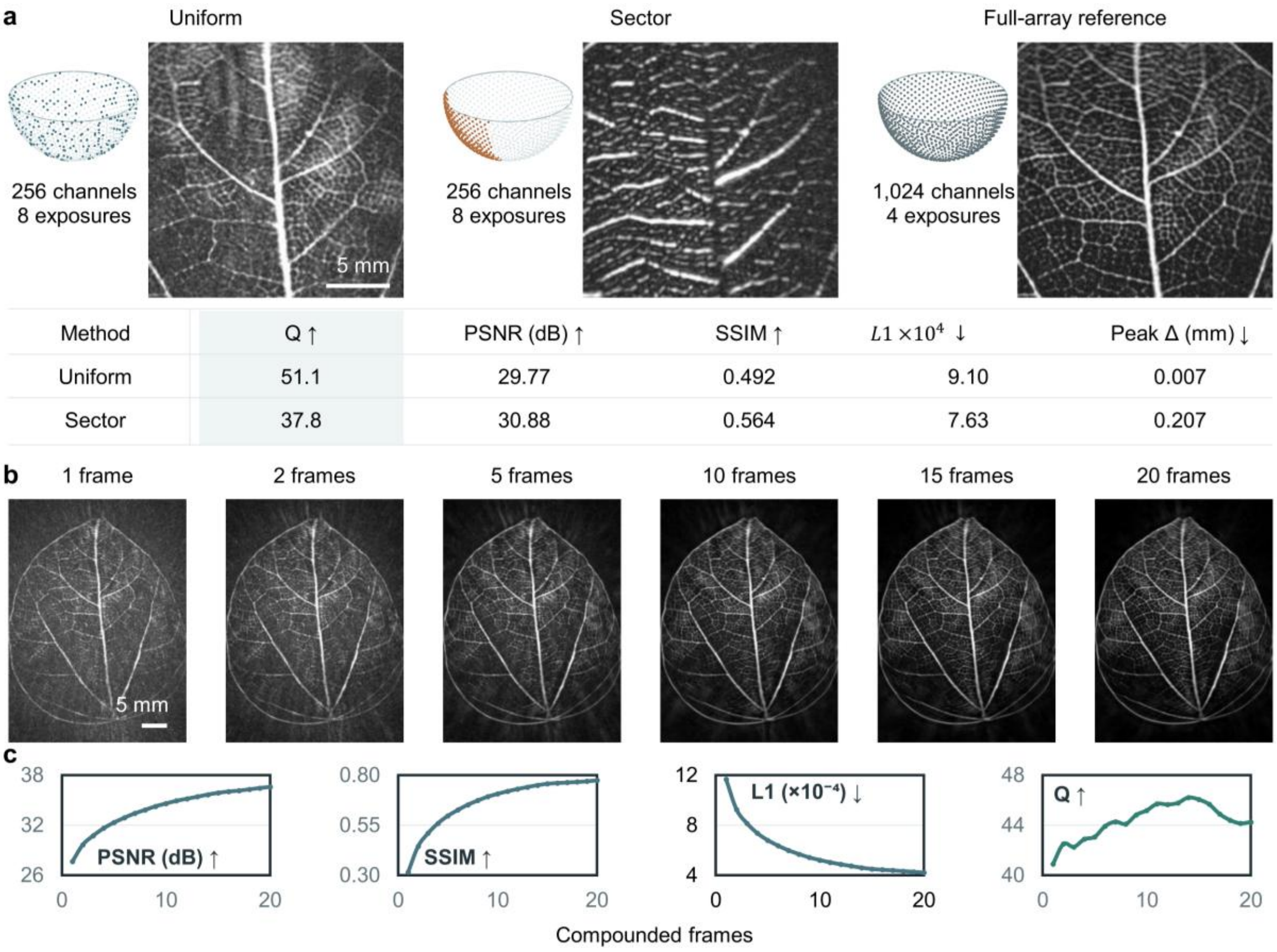

| Method | Q ↑ | PSNR (dB) ↑ | SSIM ↑ | $L1 \times 10^4$ ↓ | Peak Δ (mm) ↓ |
|---|---|---|---|---|---|
| Uniform | 51.1 | 29.77 | 0.492 | 9.10 | 0.007 |
| Sector | 37.8 | 30.88 | 0.564 | 7.63 | 0.207 |

**Figure 7. Angular sampling and exposure compounding in a measured leaf.** (a) Uniform and sector sampling, each using 256 channels and eight exposures. The displayed full-array image and scoring reference use separate exposure sets (Section 2.4). Peak Δ is the median absolute displacement over 40 reference profiles. (b) Selected rotational compounds with a shared linear display window. (c) Metrics for 1–20 exposures against the disjoint 21–40-exposure reference. Scale bars, 5 mm.

Whole-volume metrics favor the sector: PSNR is 30.88 versus 29.77 dB, SSIM is 0.564 versus 0.492, and L1 is lower. The recorded selections link these differing responses to a defined sampling comparison. Peak differences measure localization agreement along selected profiles from this acquisition, rather than imaging resolution.

Nested rotational compounds use the same acquisition to examine exposure budget (Fig. 7b,c). From one to twenty exposures, PSNR increases from 27.65 to 36.59 dB, SSIM from 0.316 to 0.774, and L1 decreases from $1.166 \times 10^{-3}$ to $4.210 \times 10^{-4}$. Q rises from 40.9 to 46.2 at fourteen exposures and is 44.3 at twenty exposures. Thus, accumulating measurements improves whole-volume reference agreement while local-contrast agreement follows a different trajectory. The retained exposure selections and disjoint reference allow users to examine this behavior in the numerical volumes. The Q maximum describes this acquisition and reference, with dependence among the nested exposure subsets.

Finally, the stored BP and MB volumes extend the assessment to different reconstruction routes, with processing records supporting inspection and rescoring (Fig. 8 and Table 2).

After independent per-volume normalization, five-exposure BP has higher Q, PSNR and SSIM and lower L1 than single-exposure BP. For the stored five-iteration MB outputs, five exposures increase Q and PSNR and reduce L1, while SSIM decreases slightly. Orthogonal MIPs and matched details show the spatial differences associated with these scores.

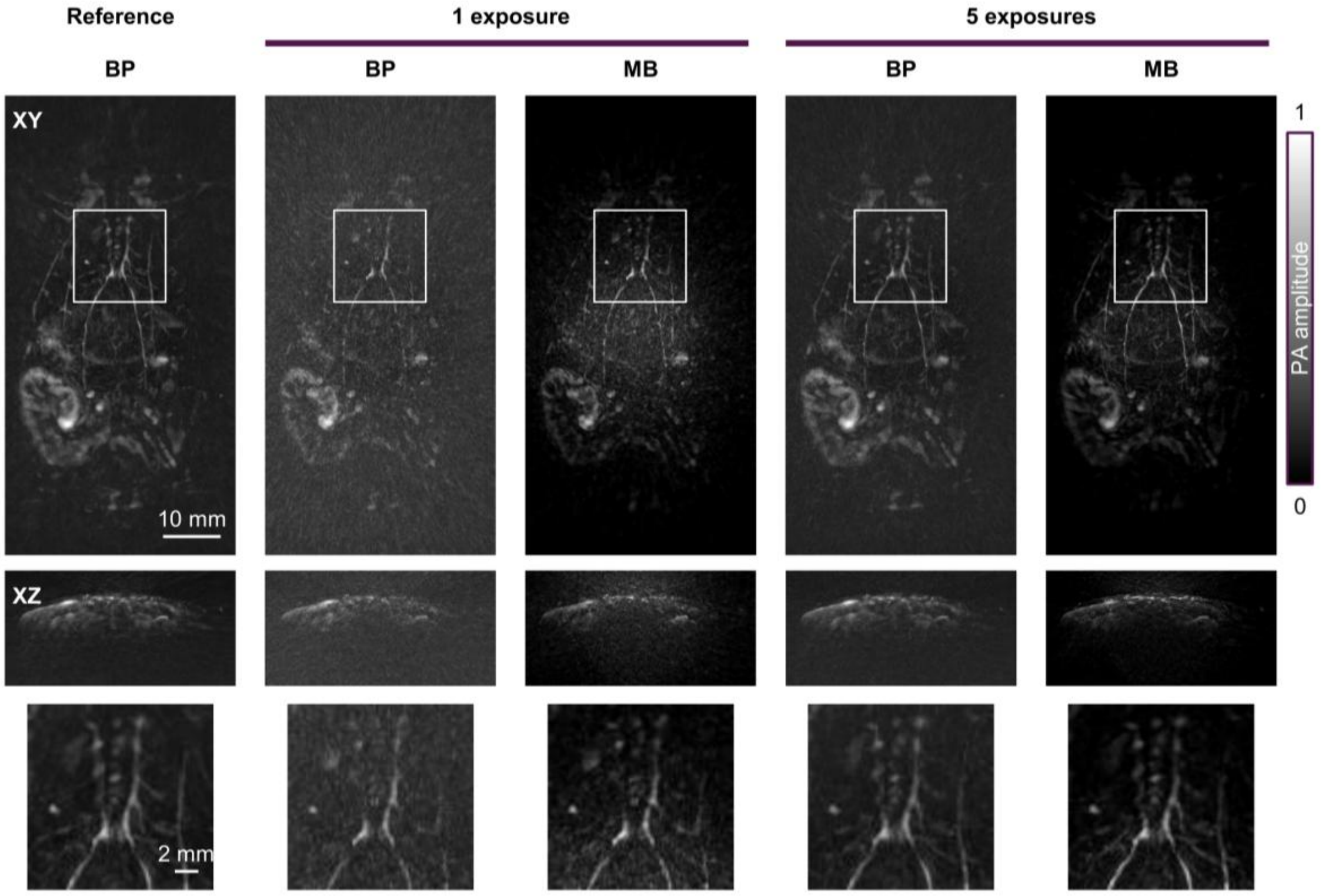

**Figure 8. BP and MB reconstructions under a shared reference.** Columns show the 50-exposure BP reference and one- and five-exposure BP and MB outputs. Rows show XY MIPs, XZ MIPs and matched details. Each volume is independently normalized to 0–1 for display and scoring. Quantitative metrics are reported in Table 2. Full-field and detail scale bars, 10 and 2 mm.

**Table 2. Quantitative comparison of independently normalized BP and MB volumes.**

| **Method** | **Exposures** | **Q ↑** | **PSNR (dB) ↑** | **Whole-volume SSIM ↑** | **L1 (×10⁻²) ↓** |
|---|---|---|---|---|---|
| BP | 1 | 63.5 | 25.97 | 0.345 | 3.84 |
| MB | 1 | 44.3 | 24.90 | 0.134 | 4.79 |
| BP | 5 | 83.8 | 33.16 | 0.770 | 1.66 |
| MB | 5 | 53.4 | 25.00 | 0.122 | 4.74 |

Reference: 50-exposure BP. Each volume is independently min–max normalized to [0, 1] and scored on its native ZYX [101, 401, 201] array. SSIM includes background and is not restricted to the Q mask. L1 values are in units of $10^{-2}$; 3.84 corresponds to 0.0384. MB uses five fixed iterations. ↑/↓ indicate higher/lower values are better. The reference contains the candidate exposures; scores describe normalized reference agreement.

The retained records specify the conditions needed to interpret the comparison. The 50-exposure BP reference contains the candidate inputs, and each reconstruction branch uses its recorded preprocessing and iteration settings. Scores therefore describe agreement with that reference and cannot independently establish algorithmic accuracy or absolute amplitude fidelity. Recovered MB parameters and objective histories document the stored iterative outputs (Supplementary Methods S5). Together with the input selections and numerical volumes, these records support reanalysis of a comparison whose reference and processing choices remain explicit.

## 4 Discussion

PACTHub addresses a data-resource need in three-dimensional PACT: access to substantial experimental collections whose measurements can be revisited across different specimen categories and acquisition modes. Existing numerical, cross-sectional, quantitative-phantom and volumetric resources support important research questions [1,2,4–18,33,34]. PACTHub adds an archive of 1,464 acquisition groups and 99,810 exposures organized for repeated reconstruction and analysis. Its contribution is the combination of measured data, acquisition coverage and a consistent representation of the physical information required for reuse.

The scale and structure of the collection have distinct roles. Multiple acquisitions provide measured variation across specimens and protocols, while repeated exposures retain alternatives within an experiment. Structured records preserve the meaning of that variation by identifying which signals, detector positions, processing choices and spatial coordinates contribute to each output. The spatial and temporal examples demonstrate the value of retaining these relationships: measurements can be combined into larger fields, followed over time or selected to change the acquisition budget. Reconstruction tools and parameter-selection procedures make these uses accessible through a common input representation, building on established efforts in photoacoustic data organization and analysis [19,20].

Evaluation should retain the same attention to physical and numerical context. In sparse PACT volumes, small structures and background occupy very different fractions of the field. The phantom and leaf examples show that whole-volume agreement and reference-supported local contrast can respond differently to a reconstruction change. Q contributes the latter readout alongside PSNR, SSIM and L1. Its support-external false-vessel example also exposes the need to inspect the rest of the volume. Providing the candidate selections, reference definitions and numerical outputs lets users examine such differences within the recorded experiment.

These properties make PACTHub useful as a data foundation for AI research. Prior PACT studies demonstrate learning-based reconstruction and sparse-data enhancement [22,23]. The records support reduced-measurement reconstruction, view aggregation, spatial fusion and temporal analysis, with inputs derived from documented acquisition subsets. Training, validation and test sets should be split by acquisition day, with all records acquired on the same day assigned to the same partition. This prevents sample-level leakage caused by distributing related exposures, scan positions or time points across the sets. Known same-specimen and duplicate relationships must also be preserved across dates. Held-out days then support evaluation on separate acquisition sessions, while subject-disjoint evaluation requires the corresponding identity information.

The archive has limits that define its present uses. Acquisition relationships are retained, but subject independence cannot be verified across the collection; subject-disjoint evaluation therefore requires additional identity information. The temporal analyses use exploratory ROI selection and report relative signals, with physiological interpretation dependent on experimental context. Reconstruction comparisons remain sensitive to reference construction, preprocessing and normalization. Public UBP permits reconstruction through the shared interface, while exact regeneration of historical in-house BP outputs remains distinct from rescoring stored volumes.

PACTHub will be maintained as a long-term data resource. Additional experimental measurements and data acquired with further detector-array configurations will be curated and progressively made public through the same PAdata/JSON structure, retaining array-specific geometry, acquisition conditions and reconstruction settings. This expansion will provide a basis for evaluating methods across acquisition conditions and array designs.

## 5 Conclusions

PACTHub expands access to experimental three-dimensional PACT data through a structured collection of 1,464 acquisition groups and 99,810 exposures. Animal, phantom and human measurements span single-position, rotational, panoramic and temporal acquisition modes. Linking recorded signals to physical reconstruction settings and numerical outputs preserves the information needed to revisit image formation and analysis. Spatial, temporal and sampling examples show how the same records support multiple investigations through a common data representation and reconstruction workflow. The combination of acquisition scale, experimental coverage and structured access addresses a gap in reusable volumetric

PACT resources. It provides an experimental data foundation for developing and evaluating AI methods. Long-term maintenance and the progressive addition of measurements from further array configurations will extend this foundation for reusable three-dimensional PACT research and AI development.

## Data availability

The full date-organized PACTHub collection is available through Google Drive at https://drive.google.com/drive/folders/1zBrduTKe5gnsNmlOfsDCgB9y5JTp5x3b. It contains PAdata and associated reconstruction parameters. Published reference volumes cover 1,315 acquisition groups and have 1,525 native-resolution XY MIP previews, comprising 1,315 GT20 and 210 Scan Reference projections. An interactive visual preview of the dataset is provided at https://pacthub.net/, together with documentation and access links. A representative subset is archived at https://doi.org/10.5281/zenodo.23496053 under CC BY 4.0. The PACTHub inventory and exclusion list accompany the article source data. The animal overview reuses the author-released processed 3D-PanoPACT volume [2]. Supplementary Methods S7 describes the source-file associations. Access to additional data and visual previews will continue to be provided through pacthub.net.

## Code availability

The public reconstruction workflow is available at https://github.com/ZhiboXiao/PACTHub. It provides the PAdata/JSON interface, FOV and effective sound-speed selection, literature-based UBP [29], and an MB solver with finite-disk forward and adjoint operators. The in-house BP core used for the stored comparison volumes is confidential; only its pseudocode is provided (Supplementary Methods S6). Accompanying article source files are described in Supplementary Methods S7. They comprise the Q implementation, digital phantom arrays and generator, temporal measurements and ROI records, leaf selection records, stored-output checksums, and the public-UBP leaf profile and numerical result. The source manifest records file hashes and distinguishes numerical reconstruction from display processing.

## Ethics approval and consent

Animal experiments were approved by the Department of Science and Technology of Zhejiang Province. Informed consent was obtained from all participants in the human experiments.

## Funding

This work was supported by the National Natural Science Foundation of China [2021MG1BI01].

## Declaration of competing interest

The authors declare no competing interests.

## Declaration of generative AI and AI-assisted technologies in the manuscript preparation process

During the preparation of this work, the authors used ChatGPT to assist with code preparation, manuscript organization and English-language editing.